%% file: ms.tex
\pdfoutput=1
\documentclass[11pt,letterpaper]{article}
\usepackage{amsmath}
\usepackage{amssymb}
\usepackage{breqn}
\usepackage[ampersand]{easylist}
\usepackage{emnlp2017}
\usepackage{times}
\usepackage{latexsym}
\usepackage{placeins}
\usepackage{graphicx}
\usepackage{bm}
\usepackage{lipsum}
\usepackage{array}
\newcolumntype{P}[1]{>{\centering\arraybackslash}p{#1}}
\newcolumntype{H}{>{\setbox0=\hbox\bgroup}c<{\egroup}@{}}

\newcommand{\secupa}{\vspace{-0.0cm}}
\newcommand{\secupb}{\vspace{-0.0cm}}
\newcommand{\ssecupa}{\vspace{-0.0cm}}
\newcommand{\ssecupb}{\vspace{-0.0cm}}
\newcommand{\titlemoveup}{\vspace{-0.mm}}
\newcommand{\eat}[1]{}

\emnlpfinalcopy

\def\emnlppaperid{***}

\title{Sentence Simplification: Challenges for Encoder-Decoder Models}
\title{Learning to Simplify Sentences with Sequence to Sequence Models}
\title{\titlemoveup S4: Simplifying Sentences with Sequence to Sequence Models}
\title{\titlemoveup Simplifying Sentences with Sequence to Sequence Models}

\author{Alexander Mathews$^{*\dagger}$, Lexing Xie$^{*\dagger}$, Xuming He$^\ddagger$\\
Australian National University$^*$, Data to Decision CRC$^\dagger$, ShanghaiTech University$^\ddagger$\\
{\tt\small alex.mathews@anu.edu.au, lexing.xie@anu.edu.au, hexm@shanghaitech.edu.cn}
}

\date{}

\begin{document}

\maketitle

\author{Alexander Mathews}

\input{abstract}

\input{intro}
\input{related_work}
\input{model}

\input{eval}

\input{result}

\secupa
\section{Conclusion}
\secupb
We present S4, a sequence-to-sequence model for simplifying sentences. The new loss function encourages word copying, reducing the requirements on the word generator, and thereby narrowing its to focus to the changes necessary for simplification. Word-copy feeding ensures the model sees an accurate word history even when copying is used extensively. Our tune-able method for incorporating pre-trained word embeddings into the pipeline allows efficient use of external data -- though much work in this area remains.

The remaining obstacles include low reliability of word substitutions, and the lack of aligned data.
Future work includes exploiting datasets from related tasks.

\section*{Acknowledgements}
The "Newsela dataset" used in this work was provided by Newsela.
This work is supported, in part, by the Australian Research Council via project DP180101985.
The Tesla K40 used for this research was donated by the NVIDIA Corporation.

\FloatBarrier
\vspace{0.6cm}
\bibliography{ms}
\bibliographystyle{emnlp_natbib}

\FloatBarrier
\newpage

\onecolumn
\section{Appendix}

\subsection{Generating Sentence Level Alignments for Newsela Dataset}
\label{ssec:sent_align_sup}

The newsela dataset in its raw form is aligned only at the document level. We align the different rewrites at the sentence level using a dynamic programming algorithm loosely based on the work of Coster \& Kauchak~\cite{Coster2011}. Our approach allows sentence splitting, where two simple sentences align to a single complex sentence. We also take sentence ordering into account, which permits a low similarity score threshold, which increases the number of matches.

The main components of any dynamic programming algorithm are the sub-problems, which break the task into manageable chunks, and recurrence relationship which describe how to combine solutions to sub-problems. Our sub-problem, denoted $a(i,j)$, is the optimal score for aligning all sentences in the complex document after and including index $i$ to all sentences in the simple document after and including index $j$. The recurrence relation which describes how to build up these sub-problems is defined in Equations \ref{equ:s4_single_rec}-\ref{equ:s4_full_rec}. First, we define $s_i$ as the $i$'th complex sentence and $s_j$ as the $j$'th simplified sentence. Note that in this section, $i$ and $j$ denote sentence indices rather than word indices as was the case in Section~\ref{ssec:custom_loss}. The similarity function between two sentences is denoted $d_{i,j}$ and defined in Equation~\ref{equ:s4_bleu_sim}. $D_{comp}$ is the number of sentences in the complex document, and $M_i$ is the number of words in the $i$'th complex sentence. $D_{simp}$ is the number of sentences in the simplified document, and $L_j$ is the number of words in the $j$'th simplified sentence. 

For clarity we present the recurrence relationship in two parts, the first matches each complex sentence with a single simple sentence, the second matches each complex sentence with two simplified sentences. 

For single sentence matching there are three possible cases reflected in Equation~\ref{equ:s4_single_rec}. If we choose to match the sentences the score is the similarity of the two sentences $d_{i,j}$, plus the best score for all later alignments $a(i+1, j+1)$. If we choose not to match the two sentences, the score is $\gamma$ (the skip penalty) plus the best score for all later alignments: this is $a(i+1, j)$ if we skip the complex sentence and $a(i,j+1)$ if we skip the simple sentence.

\begin{dmath}
a^{single}(i, j) = 
\max(a(i+1, j+1) + d_{i,j}, a(i+1, j) + \gamma, a(i, j+1) + \gamma)
\label{equ:s4_single_rec}
\end{dmath}

The sentence similarity function is defined in terms of the BLEU-4 score as:
\begin{dmath}
{\sigma(s_i, s_j) = \text{BLEU-4}(s_i, s_j)}\\
{d_{i, j} = \min(\sigma(s_i, s_j), \sigma(s_j, s_i))}
\label{equ:s4_bleu_sim}
\end{dmath}
We use BLEU-4 because its sensitivity extends up to a four-gram overlap, but also includes tri-gram, bi-gram and uni-gram overlap. We found BLEU-4 gave reasonable alignments in most cases.
Note that BLEU-4 varies between 0 and 100, with 100 being the highest similarity.

For multi-sentence matches we consider splitting the complex sentence into two parts. The recurrence is described in Equation \ref{equ:s4_multi_rec}. Here $p$ is the split index for the complex sentence, with each fragment aligned to a different simplified sentence. Since we consider local sentence re-ordering there are two options for each index $p$. The first option has the complex sentence prefix aligned to the first simple sentence and the suffix aligned to the second simple sentence. The second option has the prefix aligned to the second sentence and the suffix aligned to the first sentence. In both cases we add the best score for all later alignments $a(i+1, j+2)$.

\begin{dmath}
a^{multi}(i,j)=\max_{p<D_{comp}}\max({\sigma(s_{i,[1:p]}, s_j) + \sigma(s_{i,[p:M]}, s_{j+1}) },\\ \sigma(s_{i,[1:p]}, s_{j+1}) + \sigma(s_{i,[p:M]}, s_{j})) + a(i+1, j+2)
\label{equ:s4_multi_rec}
\end{dmath}

The notation $[\alpha:\beta]$ denotes all integer values between $\alpha$ and $\beta$, inclusive. Using Equation~\ref{equ:s4_multi_rec} and Equation~\ref{equ:s4_single_rec} we define the full recurrence as:

\begin{dmath}
a(i, j) = 
\begin{cases}
0 & \text{if}\ i \ge D_{simp}\\
0 & \text{if}\ j \ge D_{comp}\\
\max(a^{single}(i, j), a^{multi}(i,j)) & \text{otherwise}
\end{cases}
\label{equ:s4_full_rec}
\end{dmath}

As is usual in dynamic programming algorithm this recurrence is efficiently computable when caching sub-problems. Once the optimal alignment score is found the alignments can be recovered by backtracking through the cache and choosing the action at each $i$, $j$ position that lead to the optimal score.

\subsection{Overall results: Pre-trained Word-vectors}
\label{ssec:pre_train_grid_sup}

\begin{figure}
\begin{center}
\includegraphics[width=0.8\textwidth]{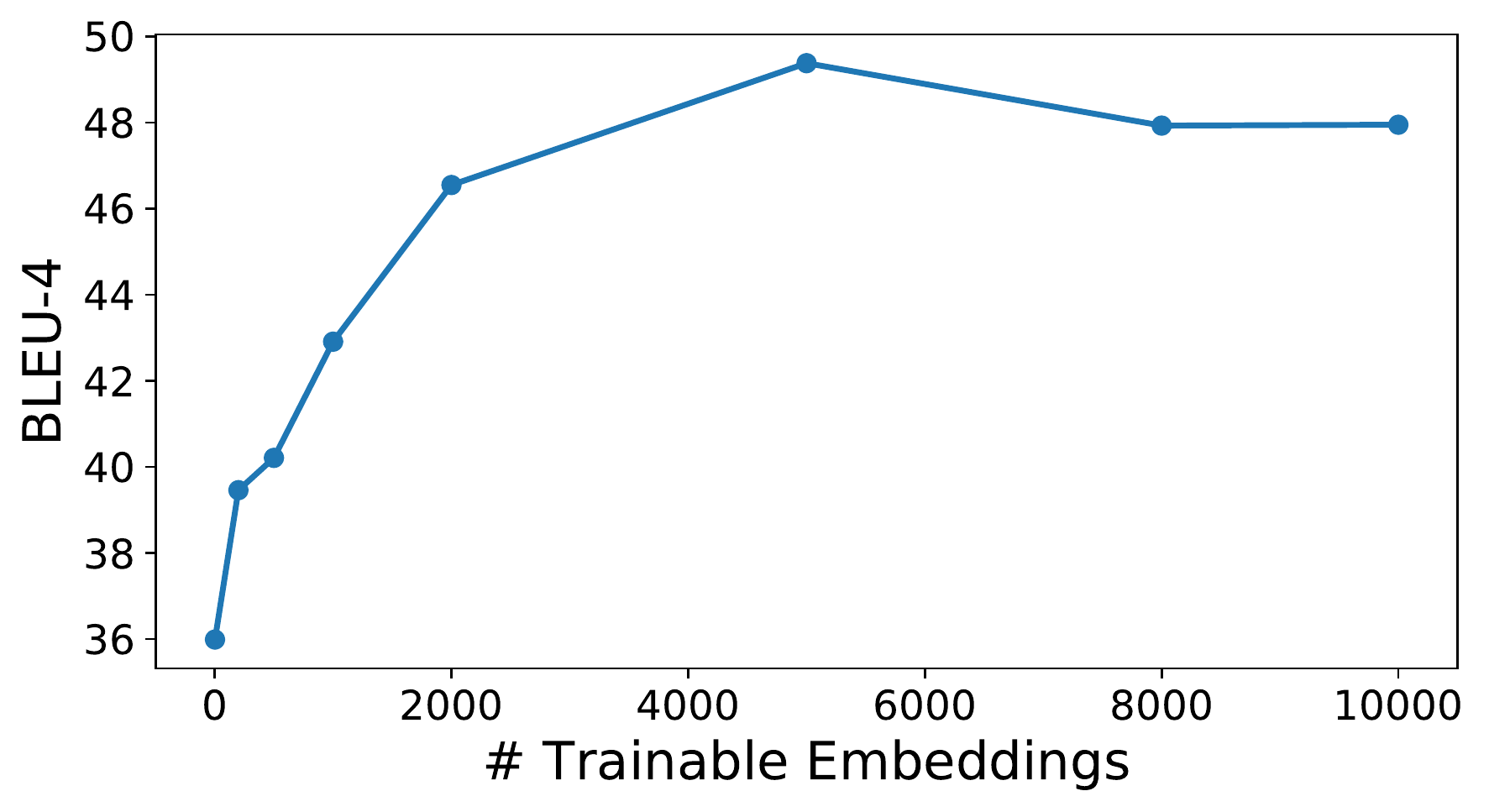}
\caption{The validation performance of \textit{S4+gv+bce} with different numbers of trainable and pre-trained embeddings. Higher BLEU-4 scores indicate greater simplification precision.}
\label{fig:s4_emb_vs_bleu}
\end{center}
\end{figure}

By extending the vocabulary with pre-trained word-vectors (\textit{S4+gv}) we mitigate the affects of data scarcity.  Figure~\ref{fig:s4_emb_vs_bleu} and Table~\ref{tab:change_emb_vocab_size} shows that using 5000 trainable embeddings gives the best validation performance. If instead we use too many fixed embeddings (left of Figure~\ref{fig:s4_emb_vs_bleu}) the model lacks the flexibility necessary to learn accurate simplifications, likewise too many trainable embeddings means the model lacks the information to deal with uncommon words (right of Figure~\ref{fig:s4_emb_vs_bleu}). This is a classic bias vs variance trade-off.

\begin{table*}
\begin{tabular}{|c||c|c|c|c|c|c|c|c|}
	\hline
	\bf{Trainable Size}& \bf{B-1} & \bf{B-2} & \bf{B-3} & \bf{B-4} & \bf{Rouge} & \bf{Flesch} & \bf{Avg.Words} & \bf{Edit Dist.}\\ \hline\hline
	\bf{2} & 0.6030 & 0.4943 & 0.4193 & 0.3599 & 0.6294 & - & 15.56 & 6.3734\\ \hline
	\bf{200} & 0.6241 & 0.5236 & 0.4524 & 0.3946 & 0.6513 & 77.6494 & 15.6148 & 5.5246 \\ \hline
	\bf{500} & 0.6303 & 0.5313 & 0.4602 & 0.4021 & 0.6549 & 77.0174 & 15.7187 & 5.3652 \\ \hline
	\bf{1000} & 0.6480 & 0.5527 & 0.4849 & 0.4291 & 0.6719 & 76.2988 & 15.7867 & 5.1207 \\ \hline
	\bf{2000} & 0.6689 & 0.5813 & 0.5182 & 0.4655 & 0.6934 & 75.2090 & 15.8332 & 4.5082 \\ \hline
	\bf{5000} & 0.6827 & 0.6012 & 0.5427 & 0.4938 & 0.7073 & 73.8551 & 15.8519 & 3.9816 \\ \hline
	\bf{8000} & 0.6709 & 0.5876 & 0.5284 & 0.4793 & 0.6987 & 75.4536 & 15.724 & 4.6011 \\ \hline
	\bf{10000} & 0.6736 & 0.5891 & 0.5291 & 0.4795 & 0.6987 & 75.1255 & 15.9078 & 4.4410 \\ \hline

\end{tabular}

\caption{Changing the number of trainable embeddings. All non-trainable embeddings are fixed to pre-trained GloVe vectors.}
\label{tab:change_emb_vocab_size}
\end{table*}

\subsection{Word Replacement Performance}
\label{ssec:word_replace_sup}
We examine the performance of word replacement in isolation from the attention by using the hand-aligned data (Section~\ref{sec:dataset}). The goal of the decoder becomes computing the most likely next word given: the original sentence, all previously generated words and the known alignment for the next word. To incorporate ground-truth alignments into the model we replace the attention $a_{i,j}$ with the count normalised ground-truth alignments $\hat{a}_{i,j} = \frac{a^{gt}_{i, j}}{\sum_{m=0}^{L}{a^{gt}_{i, m}} } $. Where $a^{gt}_{i,j}$ is a ground-truth alignment indicator, with unit value if the $j$'th input word is aligned to the $i$'th output word, zero otherwise. Normalisation is necessary because multi-word alignments are permitted, and occur frequently in practice.

\begin{table}
\begin{center}
\begin{tabular}{|c||c|c|}
\hline  \textbf{Generated} & \multicolumn{2}{c|}{\textbf{Ground Truth}} \\
\hline  & Copy Word & Change Word \\ 
\hline \hline Copy Word & 5359 & 259 \\ 
\hline Change Word & 227 & 106 \\ 
\hline 
\end{tabular}
\caption{Confusion matrix for choosing to change or copy a word. The rows are the actions chosen by the {\em S4+gv+bce} model when fed ground-truth alignments. The columns are the ground truth actions.}
\label{tab:s4_changed_words}
\end{center}
\end{table}

Table \ref{tab:s4_changed_words} is the confusion matrix for \textit{S4+gv+bce}, showing changed words (words undergoing substitution) and copied words (words copied directly from the input). \textit{S4+gv+bce} frequently chooses to copy words, mirroring the high similarity between ground-truth sentences. However, in only 32\% of cases does \textit{S4+gv+bce} correctly choose to change a word. Even when \textit{S4+gv+bce} correctly decides to change a word, it only chooses the same word as the ground-truth 46\% of the time. Word replacement itself does not perform well. Better word replacement may be archived with significantly more training data, though new ideas seem necessary for further improvement in the more common case of data scarcity.

\newpage

\end{document}

%% file: abstract.tex
\begin{abstract}

We simplify sentences with an attentive neural network sequence to sequence model, dubbed S4.
The model includes a novel word-copy mechanism and loss function to exploit linguistic similarities between the original and simplified sentences. It also jointly uses pre-trained and fine-tuned word embeddings to capture the semantics of complex sentences and to mitigate the effects of limited data.
When trained and evaluated on pairs of sentences from thousands of news articles, we observe a 8.8 point improvement in BLEU score over a sequence to sequence baseline; however, learning word substitutions remains difficult. Such sequence to sequence models are promising for other text generation tasks such as style transfer.

\end{abstract}

%% file: intro.tex
\secupa
\section{Introduction}
\secupb
Texts come in different levels of complexity, from technical pieces written for domain experts to simple books for children. Automated text simplification -- rewriting to make the language easier to understand while preserving semantics -- would allow one to more quickly digest information outside one's specific background. The benefits would be even greater for new language learners or people with language impairments. It could adapt complex texts for a large audience, reduce misinformation and aid information flow between different cultures and technical disciplines. This work tackles an important sub-problem -- text simplification by sentence rewriting.

Our goal is to learn end-to-end sentence level simplification using a parallel corpora. Sentence simplification is a complex problem and there have been many attempts to solve it using phrase translation~\cite{Coster2011,Wubben2012,Stajner2015}, parse tree translation~\cite{Cohn2008, Zhu2010}, external paraphrase corpora~\cite{Xu2016b,Napoles2016}, or by reducing the problem to lexical simplification~\cite{Specia2012, Paetzold2015, Horn2014}. Much of this work uses simple wikipedia~\cite{Zhu2010, Coster2011, Horn2014}; however, Xu et al.~\shortcite{Xu2015a} recently showed that this data contains a large number of inadequate simplifications and is prone to sentence alignment errors. We chose to focus on learning sentence simplification, by bringing together a new parallel corpus at the sentence level~\cite{Xu2015a}, and successful sequence to sequence models for machine translation~\cite{Wu2016d,Sutskever2014,Luong2015a} and sentence compression~\cite{Filippova2015, Rush2015}. 
 
We adapt neural {\em S}equence to {\em S}equence models~\cite{Sutskever2014,Luong2015a} for the {\em S}entence {\em S}implification problem, dubbed S4.
We incorporate a large vocabulary with both pre-trained and learned word embeddings to 
mitigate the effects of limited training dataset. Next, we develop a novel word-copy feeding algorithm that, with the aid of attention, exploits linguistic similarities between the original and simplified sentences.  We then introduce a novel loss function to encourage further word-copying, which allows the output sentence to benefit from a rich vocabulary despite having limited training data. Our generated sentences are simpler than the input and preserves the meaning of the original sentence. Compared to reference sentences, word-copying with the novel objective improves BLEU-4 by 4.9 points, and using the right mixture of pre-trained and learnt embeddings leads to a further 3.8 point improvement. 

Section \ref{sec:model} describes the attentive sequence to sequence model and novel adaptations. Section \ref{sec:dataset} discusses the dataset and evaluation, followed by results for each model component in Section \ref{sec:results}.

%% file: related_work.tex
\section{Related Work}

Section~\ref{ssec:methods} examines previous attempts at sentence simplification and the related tasks of machine translation, lexical simplification, sentence compression, and summarization. Section~\ref{ssec:datasets} explores the datasets available for sentence simplification.

\subsection{Methods}
\label{ssec:methods}
Sentence simplification sits within a set of re-writing tasks, including: machine translation, lexical simplification, sentence compression, and summarization. However, none of the aforementioned re-writing tasks are solved, nor are they drop in solutions to sentence simplification, so they can only guide our approach.

\noindent{\bf Machine translation} is similar to sentence simplification, though machine translation is a more developed area with many system designs already having being thoroughly explored. There have been attempts to adapt machine translation methods to sentence simplification, including: phrase translation~\cite{Coster2011,Wubben2012,Stajner2015}, parse tree translation~\cite{Cohn2008, Zhu2010}, and external paraphrase corpora~\cite{Xu2016b,Napoles2016}. Phrase-based machine translation is the most common technique adapted to sentence simplification~\cite{Wubben2012, Stajner2015}, in part because of open source libraries such as Moses~\cite{Koehn2007a}. \citet{Wubben2012} use Moses to generate a short list of candidates which they re-rank by levenshtein distance to the input. \citet{Stajner2015} evaluate the affect of training data size and quality on simplifications generated by Moses. Both \citet{Wubben2012} and \citet{Stajner2015} show that phrase-base machine translation outperforms some simple baselines but do not consistently outperform the unmodified input text as judged by machine translation metrics.

\noindent{\bf Lexical simplification} is a sub-problem of sentence simplification, involving the replacement of a word or phrase with a simpler alternative -- re-ordering or deletion are not permitted. The problem can be broken down, into complex word identification~\cite{Paetzold2016}, and substitution selection~\cite{Specia2012}. \citet{Paetzold2015} summarise a range of feature based approaches~\cite{Szarvas2013,Horn2014} and develop a modular toolkit named LEXenstein that tackles both subtasks. This toolkit identifies complex words with a binary classifier, selects word substitution candidates with word2vec~\cite{mikolov2013word2vec} and then re-ranks them with a binary classifier. More recent work shows embeddings from bi-directional Recurrent Neural Networks (RNNs) outperform word2vec similarity~\cite{Melamud2016} for substitution selection. The affect of RNNs on the entire lexical simplification pipeline has yet to be explored; however, this result suggests RNNs are capable of capturing the broader semantic context necessary for simplification.

\noindent{\bf Sentence compression} involves reducing the length of a sentence by removing phrases, while retaining grammatical correctness and the original meaning. This task targets short output sentences, without requiring that they are simpler than the input. Previous solutions relied on external corpora and parse trees~\cite{jing2000sentred,cohn2009trans}, more recently large parallel corpora~\cite{Filippova2013} have lead to interest in end-to-end learning~\cite{Filippova2015,Rush2015,Auli2015}. \citet{Filippova2015} use a neural network encoder-decoder model to tackle sentence compression. They train end-to-end on a parallel corpus of 2 million sentences built from news article headlines and first sentences~\cite{Filippova2013}. Their model beats the state-of-the-art approach in automatic and human evaluations. Other authors~\cite{Rush2015,Auli2015} extend this model to abstractive compression, where generated words are not a strict subset of the original sentence.

\subsection{Datasets}
\label{ssec:datasets}

Many recent attempts at sentence simplification~\cite{Zhu2010, Coster2011, Horn2014} use the simple wikipedia dataset~\cite{Zhu2010}. This dataset was constructed by aligning sentences from paired articles in English Wikipedia and Simple English Wikipedia\footnote{\url{https://simple.wikipedia.org}}. The Simple English Wikipedia is written by volunteers in a similar way to English Wikipedia, though they are encouraged to use only the 1000 most common English words, simple grammar, and shorter sentences. These are not strictly enforced, but rather considered broad guidelines. For example using words outside the 1000 most common is permitted, and relatively frequent in practice. The simple wikipedia dataset consists of 108,016 paired sentences extracted from 65,133 articles; the average sentence length is 25.01 in wikipedia and 20.87 in simple wikipedia.

\citet{Xu2015a} recently showed that simple wikipedia dataset contains a large number of inadequate simplifications and is prone to sentence alignment errors. They instead suggest the Newsela dataset, sourced (with permission) from the online news source Newsela\footnote{\url{https://newsela.com/}}, which consists of news articles re-written by professional editors to target different reading grades. These are roughly aligned with grades 3, 4, 6, 7 and 12, under the Common Core Standards in the United States. A thorough analysis by \citet{Xu2015a} shows that compared to simple wikipedia, Newsela has a more consistent level of quality with a higher degree of simplification. They estimate that only 50\% of sentences in simple wikipedia are true simplifications, while at least 90\% of Newsela sentence pairs are true simplifications. The number of true simplifications increases to 92\% when only the alignments between the most complex articles and the most simple articles are considered. We use of the Newsela dataset, because of its consistent quality and higher degree of simplification when compared to the simple wikipedia dataset.

%% file: model.tex
\secupa
\section{Model}
\label{sec:model}
\secupb

Section \ref{ssec:enc-dec} provides an overview of the neural sequence to sequence model and its encoder and decoder components. Which is followed by three novel components of S4 - mixing pre-trained and trainable word embeddings (Section \ref{ssec:word_emb}), word-copy feeding (Section \ref{ssec:word-copy_feeding}), and a custom loss function (Section \ref{ssec:custom_loss}).

We denote the inputs to the encoder and decoder as $\bm{x}^{enc}$ and $\bm{x}^{dec}$, the outputs words as $\bm{y}$, the attention vector for the $i$'th output token as $\bm{a}_i$, and the sequence state
vectors as $\bm{h}_j^{enc}$ and $\bm{h}_i^{dec}$. The encoder sequence is indexed by $j$; the decoder sequence by $i$. The encoder sequence length is $M$ and the decoder sequence length is $L$. We use bold-face for vectors and upper-case for matrices. 
\ssecupa
\subsection{Sequence to Sequence with Attention}
\label{ssec:enc-dec}
\ssecupb

\begin{figure}
\begin{center}
\includegraphics[width=0.40\textwidth]{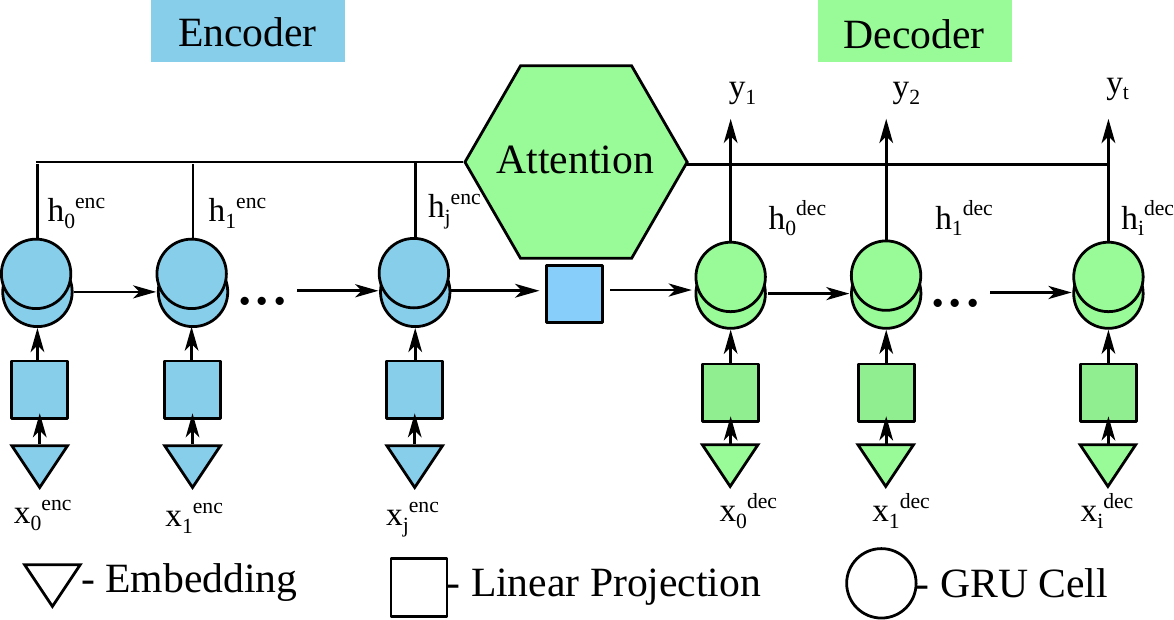}
\caption{The encoder-decoder with attention.}
\label{fig:encoder-decoder}
\end{center}
\end{figure}

Our base sequence to sequence model (Figure \ref{fig:encoder-decoder}) uses two sets of Gated Recurrent Units (GRUs)~\cite{Cho2014a}. The encoder GRU embeds the sentence it into a set of vectors, while the decoder GRU generates text from this set of vector embeddings. GRUs are a popular Recurrent Neural Network (RNN) which performs similarly~\cite{Chung2014} to the Long Short Term Memory (LSTM). The last hidden output of our encoder GRU is transformed by a fully connected linear layer and then input to the decoder GRU as the first hidden state. Both GRUs have two layers, each with 512 units, and act on sentences of up to 50 words. The 300 dimensional word embedding matrices $E^{enc}$, $E^{dec}$ are linearly projected into the 512 dimensional input space.

We implement a global attention model~\cite{Luong2015a} originally designed for machine translation. Attention is a short circuit from the sequence encoder to the sequence decoder output. In our formulation the attention vector $\bm{a}_i$ for the $i$'th output token is calculated as the softmax $\sigma(z)$ over inner products of the current decoder state with each of the encoder state vectors.
\begin{dmath}
\bm{a}_{i} = \sigma(\bm{h}_{0:M}^{enc} . (\bm{h}_i^{dec})^T)
\end{dmath}
\begin{dmath}
\sigma(\bm{z}) = \frac{e^{\bm{z}}}{\sum^M_{j=0} e^{z_j}}
\end{dmath}
The resulting attention $\bm{a}_i$ weights the output of the encoder, which forms the context vector $\bm{c}_i$.
\begin{dmath}
\bm{c}_i = \sum^M_{j=0} a_{i,j} \bm{h}_j^{enc}
\end{dmath}
The context vector is concatenated with the decoder output and input to a feed forward layer with learnt parameters $W^{out}$ and softmax non-linearity. The output is the distribution over the next word $p(y_i | x^{enc}, x^{dec}_{0:i})$.
\begin{dmath}
p(y_i | x^{enc}, x^{dec}_{0:i}) = \sigma(W^{out}[\bm{c}_{i}, \bm{h}_i^{dec}])
\end{dmath}
Where $[\bm{c}_{i}, \bm{h}_i^{dec}]$ denotes concatenation of the context and decoder hidden vectors to form a new vector.

We train end-to-end using dropout~\cite{Srivastava2014}, mini-batched adaptive gradient descent algorithm Adam~\cite{Kingma2015}, and early stopping.
Dropout was applied to the: word projection layer output, the encoder hidden outputs, the context vector, and the decoder output. The dropout ratio was set to 0.7, we found that such a large value (0.5 is more usual) helped to prevent over-fitting given our relatively small dataset and large numbers of learn-able parameters. For Adam the learning rate was set to 0.001, $\beta_1$ was 0.9 and $\beta_2$ was 0.999 -- $\beta_1, \beta_2$ are exponential decay rates for the first and second moment estimates. Note that Adam is typically insensitive to the chosen hyper-parameters~\cite{Kingma2015}. The mini-batch size was 256 sentence pairs and the score on 1024 validation samples was used for early stopping. 

\ssecupa
\subsection{Mixing Pre-trained and Trainable Word Embeddings}
\label{ssec:word_emb}
\ssecupb

A large vocabulary is necessary to represent complex sentences; however, as we show in Section~\ref{ssec:s4_ablation_study} learning embeddings for a large vocabulary when training data is limited can hurt performance. Instead we extend the size of the input vocabulary with pre-trained GloVe~\cite{Pennington2014} embeddings. Specifically, we learn embeddings for the 5000 most frequent words, and use fixed GloVe embeddings for an additional 640,317 words. The number of learnt embeddings was chosen with grid search, these results are included in appendix Section~\ref{ssec:pre_train_grid_sup}.

When the dataset covers a large range of different topics -- such as news articles -- words not seen, or infrequently seen during training may still be frequent in the test set. Pre-trained word embeddings can help to cope with this disconnect; however, using only a few learnt embeddings leads to a low variance model that cannot fit the training data effectively. By choosing a mixture of pre-trained and trainable embeddings we balance these two objectives. The learnt embeddings are restricted to the more frequent words as these have the most training data.

Extending the input vocabulary does not increase the computational cost, because we only learn embeddings for the most frequent words. Unfortunately, we cannot extend the output vocabulary without significantly increasing the computational cost of the final softmax, which is already the most expensive component for model training.

\ssecupa
\subsection{Attentive Word-Copy Feeding}
\label{ssec:word-copy_feeding}
\ssecupb

We design an attentive word-copy feeding mechanism, to copy rare words that are absent from the output vocabulary but are in the input vocabulary. This also takes advantage of the similarity between the input and output sentences.  
A special output token \textit{cpy} is introduced to denote a copy operation. When generated at position $i$, 
we copy the word $x^{enc}_{j^*}$ from the input which is the most likely alignment, computed by attention score as $j^* = \underset{j=0:M}{argmax} \{a_{i,j}\}$. With encoded sentence length denoted $M$.
This technique has been used in machine translation to deal with limited vocabulary sizes~\cite{Luong2014} but has only been applied during post-processing. In order to take advantage of a larger input vocabulary, 
we feed the copied word -- rather than the \textit{cpy} token itself -- as input $x_{i+1}^{dec}$ in the next step of sentence generation. Feeding the copied word allows the model to see more of the final sequence, which improves performance when paired with our loss function that encourages copying. This is especially important in the case of simplification where a large proportion of words are copied rather than generated as in the machine translation case.

\ssecupa
\subsection{Loss Function for Word-Copying}
\label{ssec:custom_loss}
\ssecupb
We designed a two-part loss function (Eq \ref{equ:loss}) to take advantage of the similarities between the simplified and original sentences. The first term is the categorical cross-entropy, a common loss function for encoder-decoder models~\cite{Sutskever2014,Luong2015a}, minimising it increases the probability of generating the ground truth word $\hat{y_i}$ from the softmax output. The second term is a {\em b}inary {\em c}ross-{\em e}ntropy (Eq \ref{equ:bce}). It encourages word copying at each position $i$ 
when the input word with the maximum attention $x^{enc}_{j^*}$ 
agrees with the correct ground-truth word $\hat{y}_{i}$. Intuitively, the model learns when direct copying of the input is appropriate.
We first train with the categorical cross-entropy and then fine-tune with the two-part loss function. Where $\mathbb{I}$ denotes the indicator function.

\begin{dmath}
\mathcal{L} = \frac{1}{L} ({-\sum_{i=0}^L \log P(y_i = \hat{y}_i)} + bce(\hat{y}, y))
\label{equ:loss}
\end{dmath}

\begin{dmath}
bce(\hat{y}, y) = {\sum_{i=0}^L -\log P(y_i = cpy)^{\mathbb{I}(x^{enc}_{j^*} = \hat{y}_i)}} \\
{-\log (1 - P(y_i = cpy))^{\mathbb{I}(x^{enc}_{j^*} \neq \hat{y}_i)}}
\label{equ:bce}
\end{dmath}

%% file: eval.tex
\secupa
\section{Evaluation Settings}
\label{sec:dataset}
\secupb

{\noindent\bf The Newsela dataset}~\cite{Xu2015a} is a collection of English news article sets, where each article set consists of a source article at 5 levels of simplification. The source article (considered year 12 level) was rewritten by professional linguists for (approximate) grades 3, 4, 6, 7 under the Common Core Standards in the United States. The Newsela dataset is only available upon request to Newsela, and the set of articles provided has not been standardised -- they simply provide all published articles up to the time of the request. We received 1,911 article sets, while the initial report by \shortcite{Xu2015a} consisted of 1,130 article sets. We use all 1,911 article sets to ensure maximum training data -- all our baselines use this full dataset.

The Newsela articles are grouped into sets, but the sentences are not aligned. We align sentences (as per Appendix Section~\ref{ssec:sent_align_sup}) from the 4 most complex levels to the most simple level. This choice ensures we have an approximately fixed simplification target, and a large number of aligned sentences. While other alignment options are possible we found them to be more difficult to learn.

We split the Newsela dataset into 1337 training article sets, 191 validation article sets and 383 test article sets. We remove identical aligned sentence pairs, leaving: 105,917 training sentences, 15,858 validation sentences and 28,468 test sentences. Because we split by article rather than sentence, there is a vocabulary difference between training and testing, which makes our setting more challenging.

For S4 the input vocabulary has 645,317 words. For baselines which do not exploit pre-trained embeddings the input vocabulary constructed from Newsela has 31,630 words. In all cases the output vocabulary is restricted to the most frequent 10,000 words that occur at least 7 times in training. The size of the output vocabulary is not crucial
because word copying ensures any word in the input vocabulary can end up in the simplified sentence.

{\noindent \bf Hand-alignments} To evaluate the attention mechanism, we chose a subset of 512 sentence-pairs from the validation set and created a ground-truth alignment at the word level. An automatic matching approach based on longest contiguous matching subsequence (commonly known as \textit{diff}) made an initial set of matches. One annotator (the first author) then reviewed the sentence-pairs and corrected all misaligned or missing alignments.

{\noindent \bf Evaluation metrics}

We use three types of metrics that measure: the similarity, the amount of change and the simplicity of the generated sentences. The similarity metrics BLEU~\cite{Papineni2002} (\textit{B1-B4}) and \textit{Rouge}~\cite{Lin2004} are commonly used for evaluating machine translation -- larger scores mean greater similarity to the ground-truth. The distance to the original sentence is measured by edit distance (\textit{Edit Dist.}), the number of word insertions, deletions or substitutions to turn the original sentence into the generated sentence. Sentence simplicity is measured by average words per sentence (\textit{Avg.Words}) and Flesch-Kincaid reading ease (\textit{Flesch}). \textit{Flesch} score is a widely used open-source metric for simplification tasks~\cite{Zhu2010, Narayan2014}. It weights average words per sentence and average syllables per word -- simpler sentences have higher scores.

{\noindent \bf Moses Baseline}
Many authors~\cite{Coster2011,Wubben2012,Stajner2015} have applied the open source phrase translation software \textit{Moses}~\cite{Koehn2007a} to sentence simplification. We adopt \textit{Moses} as a baseline which we train by following the directions of Coster \& Kauchak~\cite{Coster2011}. Specifically, we keep the default settings for tokenization and true-caseing, remove sentences longer than 80 words, and train a tri-gram language model using modified Kneser-Ney smoothing. The hyperparameters are tuned with Minimum Error Rate Training (MERT) which maximises the BLEU score on a sample of 400 paired sentences from the validation set. Using a small sample from the validation set is necessary because the MERT algorithm is computationally expensive; our sample size of 400 is consistent with Coster \& Kauchak.

%% file: result.tex
\secupa
\section{Results}
\label{sec:results}
\secupb

Table \ref{tab:overall_perf} summarises the performances of the model variants. The base model for \textit{S4} is an encoder-decoder model with attention and word-copy feeding. We use suffixes to show the components added or removed in each model variant: \textit{-attn} for removing attention, \textit{-feed} for removing word-copy, \textit{+gv} for allowing a mix of trainable and pre-trained embeddings, and \textit{+bce} for training with the loss function for word-copying.

\ssecupa
\subsection{Sequence to Sequence Performance}
\label{ssec:overall_perf}
\ssecupb

\begin{table*}
\begin{minipage}{0.9\textwidth}
\begin{tabular}{|l||c|c|c|c|c|c|c|c|}
	\hline
	                & \bf{B-1} & \bf{B-2} & \bf{B-3} & \bf{B-4} & \bf{Rouge} & \bf{Flesch} & \bf{Avg.Words} & \bf{Edit Dist.}\\ \hline\hline
	\bf{ground-truth} & - & - & - & - & - & 74.69 & 15.72 & 7.30 \\ \hline
	\bf{original} & 69.84 & 62.76 & 57.57 & 53.10 & 75.07 & 64.75 & 17.12 & 0.0 \\ \hline \hline
	\bf{moses} & 65.43 & 56.45 & 49.94 & 44.50 & 69.99 & 74.19 & 17.08 & 1.56 \\ \hline
	 \bf{S4-attn}   & 23.35 & 13.54 & 8.77 & 5.95 & 30.69 & 91.68 & 9.70 & 13.46\\ \hline
	 \bf{S4-feed}   & 61.94 & 51.94 & 45.14 & 39.71 & 64.75 & 75.49 & 15.49 & 5.94 \\ \hline
	 \bf{S4+gv+bce-feed} & 16.86 & 7.70 & 3.52 & 1.72 & 24.42 & 67.91 & 27.54 & 26.08 \\ \hline
	 \bf{S4}   & 63.04 & 53.60 & 46.96 & 41.51 & 65.91 & 77.90 & 15.26 & 5.53 \\ \hline
	\bf{S4+gv} & 67.51 & 59.01 & 52.90 & 47.75 & 70.28 & 73.41 & 15.34 & 3.82 \\ \hline
	\bf{S4+bce} & 65.28 & 57.23 & 51.36 & 46.43 & 68.03 & \textbf{74.72} & 15.94 & 4.70 \\ \hline
	\bf{S4+gv+bce}  & \textbf{68.71} & \textbf{60.80} & \textbf{55.11} & \textbf{50.28} & \textbf{71.02} & 68.71 & \textbf{15.80} & 3.51\\ \hline
	
\end{tabular}
\end{minipage}

\caption{Results for the end-to-end sentence simplification task. Our complete model is \textit{S4+gv+bce}. Section \ref{sec:dataset} details the metrics. Values in bold are closest to the ground-truth.}
\label{tab:overall_perf}

\end{table*}

Our model {\em S4+gv+bce} outperforms phrase translation, trained with open source software \textit{Moses}~\cite{Koehn2007a} and used by earlier simplification work~\cite{Coster2011,Wubben2012,Stajner2015}. The sentences generated by {\em S4+gv+bce} have higher BLEU and Rouge scores than \textit{Moses}, indicating greater similarity to the simplified sentences. The \textit{Moses} baseline achieves a good Flesch score; however, coupled with the lack of similarity to the simplified sentences this could indicate a loss of semantics.

We find that the \textit{original} sentences achieve high BLEU and Rouge scores despite being longer and more complex than the generated sentences -- this is consistent with previous observations~\cite{Coster2011,Wubben2012,Stajner2015}. The dataset construction is partially responsible: using BLEU-4 for aligning ground truth sentences (see Appendix Section~\ref{ssec:sent_align_sup}), means there is a large overlap between original and simplified sentences. The content domain is also responsible as there is not much change between complex sentences and those simplified by linguists. For example in the manually aligned section of the dataset only 6.1\% of aligned words were changed going from the complex to simple sentences -- the remaining 93.9\% were copied. It is also likely that the BLEU and Rouge metrics evaluation have a hand in the high performance of the \textit{original} sentences. These metrics penalise sentences shorter than the ground-truth -- BLEU explicitly and Rouge implicitly. Generated simplifications will be penalised more frequently as they tend to be shorter than the \textit{original} sentences.

\subsection{Ablation Study}
\label{ssec:s4_ablation_study}

Each component in the \textit{S4} model contributes to the performance as shown in Table~\ref{tab:overall_perf}. Removing either the attention (\textit{S4-attn}) or the feeding (\textit{S4-feed}) causes a drop in BLEU and Rouge, indicating that the generated sentences are further from the simplified ground-truth. The attention is the more important of the two, with removal leading to an enormous 35.56 BLEU-4 point drop -- we also observed semantic divergence from the input sentence after three to four words. Attention not only enables word copying but also reduces the effective depth of the network and avoids compressing the entire encoded sentence into a single fixed size vector. Adding either the pre-trained word-vectors or the custom loss function improves the BLEU and Rouge scores. 
\textit{S4+bce} has a higher BLEU-4 by 4.9 points, and \textit{S4+gv+bce} leads to a further 3.8 increase. 
Compared to the ground-truth sentences, \textit{S4+gv+bce} is closest in sentence lengths, 
while \textit{S4+bce} is closest in Flesch score. 

The custom loss and word-copy feeding are designed to be paired. 
If we keep the custom loss function but remove feeding (\textit{S4+gv+bce-feed}), performance degrades drastically, 
because the \textit{cpy} token, which provides little information by itself, is used frequently -- 87.7\% (up from 9.0\%).

\ssecupa
\subsection{Simplification Examples}
\ssecupb
Table \ref{tab:examples} shows two examples from the \textit{S4+gv+bce} model. 
The first has "massive" changed to "huge", which is semantically correct and subjectively simpler, and "grain" change to "vegetable", syntactically correct but a semantic mistake. The second example correctly splits a noun clause into two sentences. But the ground truth sentence uses additional information from article context, and is not shorter.
This demonstrates several simplification operations S4 can learn, 
it also illustrates the limitations of the available ground-truth. Having only one ground-truth simplification per original sentence can lead to penalising correctly simplified outputs. Moreover, the dataset was constructed with additional article level, and local context not provided to the sentence simplifier. 

\begin{table}
{\small
\begin{tabular}{|P{0.03\textwidth}|p{0.39\textwidth}|}
\hline Orig  & the increase would put massive strains on the world s water and grain supplies.\\ 
\hline Simp & the increase would put \textbf{huge} strains on the world s water and \textbf{vegetable} supplies.\\ 
\hline GT & the increase would strain the world s water and grain supplies. \\
\hline \hline Orig  & obama, who has not said when he ll make a \textbf{final} decision, is under heavy pressure to approve the project.\\ 
\hline Simp & obama \textbf{has} not said when he ll make a decision\textbf{. he} is under heavy pressure to approve the project.\\ 
\hline GT & president obama has not said when he ll decide what to do about the pipeline. but, he is under pressure to say yes to the project. \\
\hline
\end{tabular}
}

\caption{Simplification examples from the \textit{S4+gv+bce} model. Bold-face highlights changes.}
\label{tab:examples}

\end{table}

\ssecupa
\subsection{Attention Alignment Performance}
\ssecupb
This evaluation (Table \ref{tab:att_oricle_not_aligned}) separates, the attention component that focuses on an input word from, the effects of imperfect word substitutions. We use an oracle word simplifier that outputs the correct word from the ground-truth, if the alignment implied by the argmax attention is in the hand-alignments (Section \ref{sec:dataset}) or if the word is not aligned. Any discrepancy here is due to attention alignment not agreeing with the ground-truth. 

\begin{table}
\begin{center}
\begin{tabular}{|c||c|HHc|c|}
	\hline
	                    & \bf{B-1} & \bf{B-2} & \bf{B-3} & \bf{B-4} & \bf{Rouge} \\ \hline\hline
	   \bf{S4}     & 80.91 & 74.80 & 69.63 & 65.09 & 82.81 \\ \hline
	 \bf{S4+bce}   & 79.49 & 73.04 & 67.67 & 63.01 & 81.27 \\ \hline
	 \bf{S4+gv}   & 83.16 & 77.77 & 73.02 & 68.81 & 84.77 \\ \hline
	\bf{S4+gv+bce} & 81.92 & 76.23 & 71.27 & 66.83 & 83.50 \\ \hline
\end{tabular}

\caption{Attention alignment performance with oracle word simplifier. }
\label{tab:att_oricle_not_aligned}
\end{center}

\end{table}

Compared to Table \ref{tab:overall_perf}, having a word replacement oracle boosts BLEU by 14.4 points and Rouge by 13.7 points. 
The attention component alone performs around 81 in B-1, 
and the model variants ({\em +gv}) that use a larger vocabulary perform slightly better. 
Word replacement itself does not perform well, see Section \ref{ssec:word_replace_sup} in the appendix. Better word replacement needs non-trivial amount of training data, and new ideas
seem necessary for further improving attention alignment.